# VoxML: A Visualization Modeling Language

**James Pustejovsky and Nikhil Krishnaswamy**
Brandeis University
Waltham, Massachusetts USA
{jamesp,nkrishna}@cs.brandeis.edu

**Abstract**

We present the specification for a modeling language, VoxML, which encodes semantic knowledge of real-world objects represented as three-dimensional models, and of events and attributes related to and enacted over these objects.VoxML is intended to overcome the limitations of existing 3D visual markup languages by allowing for the encoding of a broad range of semantic knowledge that can be exploited by a variety of systems and platforms, leading to multimodal simulations of real-world scenarios using conceptual objects that represent their semantic values.

## 1. Introduction

In this paper, we describe a modeling language for constructing 3D visualizations of concepts denoted by natural language expressions. This language, VoxML (Visual Object Concept Modeling Language), is being used as the platform for creating *multimodal semantic simulations* in the context of human-computer communication.[1]

Prior work in visualization from natural language has largely focused on object placement and orientation in static scenes (Coyne and Sproat, 2001; Siskind, 2001; Chang et al., 2015), and we have endeavored to incorporate dynamic semantics and motion language into our model. In previous work (Pustejovsky and Krishnaswamy, 2014; Pustejovsky, 2013), we introduced a method for modeling natural language expressions within a 3D simulation environment built on top of the game development platform Unity (Goldstone, 2009). The goal of that work was to evaluate, through explicit visualizations of linguistic input, the semantic presuppositions inherent in the different lexical choices of an utterance. This work led to two additional lines of research: an explicit encoding for how an object is itself situated relative to its environment; and an operational characterization of how an object changes its location or how an agent acts on an object over time. The former has developed into a semantic notion of situational context, called a *habitat* (Pustejovsky, 2013; McDonald and Pustejovsky, 2014), while the latter is addressed by dynamic interpretations of event structure (Pustejovsky and Moszkowicz, 2011; Pustejovsky, 2013).

The requirements on a visual simulation include, but are not limited to, the following components:

1. A minimal embedding space (MES) for the simulation must be determined. This is the 3D region within which the state is configured or the event unfolds;

2. Object-based attributes for participants in a situation or event need to be specified; e.g., orientation, relative size, default position or pose, etc.;

3. An epistemic condition on the object and event rendering, imposing an implicit point of view (POV);

4. Agent-dependent embodiment; this determines the relative scaling of an agent and its event participants and their surroundings, as it engages in the environment.

In order to construct a robust simulation from linguistic input, an event and its participants must be embedded within an appropriate minimal embedding space. This must sufficiently enclose the event localization, while optionally including space enough for a frame of reference for the event (the viewer's perspective). We return to this issue later in the paper when constructing our simulation from the semantic interpretation associated with motion events.

Existing representation languages for 3D modeling, such as VRML (Parisi and Pesce, 1994), (Carson et al., 1999) or X3D (Brutzman and Daly, 2010), adequately represent the vertices, edges, faces, and UV texture mapping that make up the model itself, but contain no information about how such an object interacts with other objects in a real or simulated environment. Such information is represented on an *ad hoc* basis for the needs of the particular platform or environment in which the model will be deployed.

Our goal in developing VoxML is twofold: to specify a language for both designing and representing data structures that generate simulations of linguistically represented objects, properties, and events. Just as the lexical items in a language are specified and encoded with a richly typed framework, such as Generative Lexicon, most lexemes will have a representation within VoxML that encodes an interpretation of the word's semantic content as a visualization.

We have followed a strict methodology of specification development, as adopted by ISO TC37/SC4 and outlined in (Bunt, 2010) and (Ide and Romary, 2004), and as implemented with the development of ISO-TimeML (Pustejovsky et al., 2005; Pustejovsky et al., 2010) and others in the family of SemAF standards. Further, our work shares many of the goals pursued in (Dobnik et al., 2013; Dobnik and Cooper, 2013), for specifying a rigidly-defined type system for spatial representations associated with linguistic expressions.

[1]This work is being carried out in the context of CwC, a DARPA effort to identify and construct computational semantic elements, for the purpose of carrying out joint plans between a human and computer through NL discourse.

In this paper, we describe a specification language for modeling "visual object concepts", VoxML. The object defined within VoxML will be called a *voxeme*, and the library of voxemes, a *voxicon*.

## 2. Habitats and Affordances

Before we introduce the VoxML specification, we review our assumptions regarding the semantics underlying the model. Following Generative Lexicon (GL) (Pustejovsky, 1995), lexical entries in the object language are given a feature structure consisting of a word's basic type, its parameter listing, its event typing, and its qualia structure. The semantics of an object will consist of the following:

(1) a. **Atomic Structure** (FORMAL): objects expressed as basic nominal types
b. **Subatomic Structure** (CONST): mereotopological structure of objects
c. **Event Structure** (TELIC and AGENTIVE): origin and functions associated with an object
d. **Macro Object Structure**: how objects fit together in space and through coordinated activities.

Objects can be partially contextualized through their qualia structure: a food item has a TELIC value of $eat$, an instrument for writing, a TELIC of $write$, a cup, a TELIC of $hold$, and so forth. For example, the lexical semantics for the noun *chair* carries a TELIC value of $sit_in$:

$$(2)\ \lambda x \exists y \begin{bmatrix} \textbf{chair} \\ \text{AS} = \begin{bmatrix} \text{ARG1} = x : e \end{bmatrix} \\ \text{QS} = \begin{bmatrix} \text{F} = phys(x) \\ \text{T} = \lambda z, e[sit_in(e, z, x)] \end{bmatrix} \end{bmatrix}$$

While an artifact is designed for a specific purpose (its TELIC role), this can only be achieved under specific circumstances. (Pustejovsky, 2013) introduces the notion of an object's *habitat*, which encodes these circumstances. Assume that, for an artifact, $x$, given the appropriate context $\mathcal{C}$, performing the action $\pi$ will result in the intended or desired resulting state, $\mathcal{R}$, i.e., $\mathcal{C} \rightarrow [\pi]\mathcal{R}$. That is, if a context $\mathcal{C}$ (a set of contextual factors) is satisfied, then every time the activity of $\pi$ is performed, the resulting state $\mathcal{R}$ will occur. The precondition context $\mathcal{C}$ is necessary to specify, since this enables the local modality to be satisfied.
Using this notion, we define a habitat as a representation of an object situated within a partial minimal model; it is a directed enhancement of the qualia structure. Multi-dimensional affordances determine how habitats are deployed and how they modify or augment the context, and compositional operations include procedural (simulation) and operational (selection, specification, refinement) knowledge.
The habitat for an object is built by first placing it within an *embedding space* and then contextualizing it. For example, in order to use a table, the top has to be oriented upward, the surface must be accessible, and so on. A chair must also be oriented up, the seat must be free and accessible, it must be able to support the user, etc. An illustration of what the resulting knowledge structure for the habitat of a chair is shown below.

$$\lambda x \begin{bmatrix} \textbf{chair}_{hab} \\ \text{F} = [phys(x), on(x, y_1), in(x, y_2), orient(x, up)] \\ \text{C} = [seat(x_1), back(x_2), legs(x_3), clear(x_1)] \\ \text{T} = \lambda z \lambda e[\mathcal{C} \rightarrow [sit(e, z, x)]\mathcal{R}_{sit}(x)] \\ \text{A} = [made(e', w, x)] \end{bmatrix}$$

As described in more detail below, event simulations are constructed from the composition of object habitats, along with particular constraints imposed by the dynamic event structure inherent in the verb itself, when interpreted as a program.
The final step in contextualizing the semantics of an object is to operationalize the TELIC value in its habitat. This effectively is to identify the *affordance structure* for the object (Gibson, 1977; Gibson, 1979). The affordance structure available to an agent, when presented with an object, is the set of actions that can be performed with it. We refer to these as GIBSONIAN affordances, and they include "grasp", "move", "hold", "turn", etc. This is to distinguish them from more goal-directed, intentionally situated activities, what we call TELIC affordances.

## 3. The VoxML Specification

### 3.1. VoxML Elements

Entities modeled in VoxML can be objects, programs, or logical types. Objects are logical constants; programs are n-ary predicates that can take objects or other evaluated predicates as arguments; logical types can be divided into attributes, relations, and functions, all predicates which take objects as arguments. Attributes and relations evaluate to states, and functions evaluate to geometric regions. These entities can then compose into visualizations of natural language concepts and expressions.

### 3.2. Objects

The VoxML OBJECT is used for modeling nouns. The current set of OBJECT attributes is shown below:

| | |
|---|---|
| LEX | OBJECT's lexical information |
| TYPE | OBJECT's geometrical typing |
| HABITAT | OBJECT's habitat for actions |
| AFFORD_STR | OBJECT's affordance structure |
| EMBODIMENT | OBJECT's agent-relative embodiment |

The LEX attribute contains the subcomponents PRED, the predicate lexeme denoting the object, and TYPE, the object's type according to Generative Lexicon.
The TYPE attribute (different from LEX's TYPE subcomponent) contains information to define the object geometry in terms of primitives. HEAD is a primitive 3D shape that roughly describes the object's form (such as calling an apple an "ellipsoid"), or the form of the object's most semantically salient subpart. We ground our possible values for HEAD in, for completeness, mathematical formalism defining families of polyhedra (Grünbaum, 2003), and, for annotator's ease, common primitives found across the "corpus" of 3D artwork and 3D modeling software[2] (Giambruno, 2002). Using common 3D modeling primitives as convenience definitions provides some built-in redundancy to VoxML, as is found in NL description of structural forms. For example, a `rectangular_prism` is the same as a `parallelepiped` that has at least two defined planes of reflectional symmetry, meaning that an object whose HEAD is `rectangular_prism` could be defined two ways, an association which a reasoner can unify axiomatically. Possible values for HEAD are given below:

| | |
|---|---|
| HEAD | `prismatoid, pyramid, wedge, parallelepiped, cupola, frustum, cylindroid, ellipsoid, hemiellipsoid, bipyramid, rectangular_prism, toroid, sheet` |

It should be emphasized that these values are not intended to reflect the exact structure of a particular geometry, but rather a cognitive approximation of its shape, as is used in some image-recognition work (Goebel and Vincze, 2007).

Object subparts are enumerated in COMPONENTS. CONCAVITY can be `concave`, `flat`, or `convex` and refers to any concavity that deforms the HEAD shape. ROTATSYM, or rotational symmetry, defines any of the world's three orthogonal axes around which the object's geometry may be rotated for an interval of less than 360 degrees and retain identical form as the unrotated geometry. A sphere may be rotated at any interval around any of the three axes and retain the same form. A rectangular prism may be rotated 180 degrees around any of the three axes and retain the same shape. An object such as a ceiling fan would only have rotational symmetry around the Y axis. Reflectional symmetry, or REFLECTSYM, is defined similarly. If an object may be bisected by a plane defined by two of the world's three orthogonal axes and then reflected across that plane to obtain the same geometric form as the original object, it is considered to have reflectional symmetry across that plane. A sphere or rectangular prism has reflectional symmetry across the XY, XZ, and YZ planes. A wine bottle only has reflectional symmetry across the XY and YZ planes.

The possible values of ROTATSYM and REFLECTSYM are intended to be world-relative, *not* object-relative. That is, because we are only discussing objects when situated in a minimal embedding space (even an otherwise empty one) wherein all coordinates are given Cartesian values, the axis of rotational symmetry or plane of reflectional symmetry are those denoted in the world, not of the object. Thus, a tetrahedron—which in isolation has seven axes of rotational symmetry, no two of which are orthogonal—when placed in the MES such that it cognitively satisfies all "real-world" constraints, must be situated with one base downward (a tetrahedron placed any other way will fall over). Thus reducing the salient in-world axes of rotational symmetry to one: the world's Y-axis. When the orientation of the object is ambiguous relative to the world, the world should be assumed to provide the grounding value.

The HABITAT element defines habitats INTRINSIC to the object, regardless of what action it participates in, such as intrinsic orientations or surfaces, as well as EXTRINSIC habitats which must be satisfied for particular actions to take place. We can define intrinsic faces of an object in terms of its geometry and axes. The model of a computer monitor, when axis-aligned according to 3D modeling convention, aligns the screen with the world's Z-axis facing the direction of increasing Z values. When discussing the object "computer monitor," the lexeme "front" singles out the screen of the monitor as opposed to any other part. We can therefore correlate the lexeme with the geometrical representation by establishing an intrinsic habitat of the computer monitor of *front(+Z)*. We adopt the terminology of "alignment" of an object dimension, $d \in \{x, y, z\}$, with the dimension, $d'$, of its embedding space, $\mathcal{E}_{d'}$, as follows: $align(d, \mathcal{E}_{d'})$.

AFFORD_STR describes the set of specific actions, along with the requisite conditions, that the object may take part in. There are low-level affordances, called GIBSONIAN, which involve manipulation or maneuver-based actions (grasping, holding, lifting, touching); there are also TELIC affordances (Pustejovsky, 1995), which link directly to what goal-directed activity can be accomplished, by means of the GIBSONIAN affordances.

EMBODIMENT qualitatively describes the SCALE of the object compared to an in-world agent (typically assumed to be a human) as well as whether the object is typically MOVABLE by that agent.

### 3.3. Programs

PROGRAM is used for modeling verbs. The current set of PROGRAM attributes is shown below:

| | |
|---|---|
| LEX | PROGRAM's lexical information |
| TYPE | PROGRAM's event typing |
| EMBEDDING_SPACE | PROGRAM's embedding space as a function of the participants and their changes over time |

Just like OBJECTs, a PROGRAM's LEX attribute contains the subcomponents PRED, the predicate lexeme denoting the program, and TYPE, the program's type as given in a lexical semantic resource, e.g., its GL type.

TYPE contains the HEAD, its base form; ARGS, reference to the participants; and BODY, any subevents that are executed in the course of the program's operation. Top-level values for a PROGRAM's HEAD are given below:

| | |
|---|---|
| HEAD | `state, process, transition assignment, test` |

The type of a program as shown above is given in terms of how the visualization of the action is realized. Basic program distinctions, such as `test` versus `assignment` are included within this typology and further distinguished through subtyping.

[2] Mathematically curved surfaces such as spheres and cylinders are in fact represented, computed, and rendered as polyhedra by most modern 3D software.

### 3.4. Logical Types

Like OBJECTs and PROGRAMS, all functional type classes contain a LEX attribute and a PRED parameter that denote the lexeme related to the VoxML representation. The distinction between ATTRIBUTEs, RELATIONs, and FUNCTIONs lies mainly in differences in their TYPE structure, discussed in each subsection below.

#### 3.4.1. Attributes

Adjectival modification and predication involve reference to an ATTRIBUTE in our model, along with a specific value for that attribute. Often, but not always, this attribute is associated with a family of other attributes, structured according to some set of constraints, which we call a SCALE. The least constrained association is a conventional sortal classification, and its associated attribute family is the set of pairwise disjoint and non-overlapping sortal descriptions (non-super types). Following (Stevens, 1946; Luce et al., 1990), we will call this classification a `nominal` scale, and it is the least restrictive scale domain over which we can predicate an individual. `binary` classifications are a two-state subset of this domain. When we impose more constraints on the values of an attribute, we arrive at more structured domains. For example, by introducing a partial ordering over values, we can have transitive closure, assuming all orderings are defined. This is called an `ordinal` scale. When fixed units of distance are imposed between the elements on the ordering, we arrive at an `interval` scale. Finally, when a zero value is introduced, we have a scalar structure called a `ratio` scale.

In reality, of course, there are many more attribute categories than the four listed above, but the goal in VoxML is to use these types as the basis for an underlying cognitive classification for creating measurements from different attribute types. In other words, these scale types are models of cognitive strategies for structuring values for conceptual attributes associated with natural language expressions involving scalar values. VoxML encodes how attributes and the associated programs that change their values can be grouped into these scalar domains of measurement. As VoxML is intended to model visualizations of physical objects and programs, we should note here that we are only examining first-order attributes, and not those that require any subjective interpretation relative to their arguments (that is, VoxML is intended to model "the image is red" but not "the image is depressing"). Examples of different SCALE types follow:

| | | |
|---|---|---|
| `ordinal` | DIMENSION | *big, little, large, small, long, short* |
| `binary` | HARDNESS | *hard, soft* |
| `nominal` | COLOR | *red, green, blue* |
| `rational` | MASS | *1kg, 2kg, etc.* |
| `interval` | TEMPERATURE | *0°C, 100°C, etc.* |

VoxML also denotes an attribute's ARITY, or the relative of the attribute to the object it describes compared to other instances of the same object class. `transitive` attributes are considered to describe object qualities that require comparison to other object instances (e.g. *the small cup* vs. *the big cup*), whereas `intransitive` attributes do not require that comparison (a *red cup* is not red compared to other cups; it is objectively red—or its redness can be assessed quantitatively). Finally, every attribute must be realized as applying to an object, so attributes require an ARG, a variable representing said object and the typing thereof. This is denoted identically to the individual ARGS of VoxML PROGRAMs.

#### 3.4.2. Relations

A RELATION's type structure specifies a binary CLASS of the relation: `configuration` or `force_dynamic`, describing the nature of the relation. These classes themselves have subvalues—for configurational relations these are values from the region connection calculus (Randell et al., 1992). Also specified are the arguments participating in the relations. These, as above, are represented as typed variables.

#### 3.4.3. Functions

FUNCTIONs' typing structures take as ARG the OBJECT voxeme being computed over. REFERENT takes any subparameters of the ARG that are semantically salient to the function, such as the voxeme's HEAD. If unspecified, the entire voxeme should be assumed as the referent. MAPPING is set to a denotation of the type of transformation the function performs over the object, such as dimensionality reduction (notated as *dimension(n):n-1* for a function that takes in an object of `n` dimensions and returns a region of `n-1`). Finally, ORIENTATION provides three values: SPACE, which notes if the function is performed in `world` space or `object` space; AXIS, which notes the primary axis and direction the function exploits; and ARITY, which returns *transitive* or *intransitive* based on the boolean value of a specified input variable (`x[y]:intransitive` denotes a function that returns *intransitive* if the value of $y$ in $x$ is true). Defintions of transitive and intransitive follow those for ATTRIBUTEs.

## 4. Examples of Voxemes

In this section, we illustrate the representational capabilities of the specification by briefly presenting example voxeme entries from the current VoxML voxicon. VoxML OBJECT representations are intended to correspond with specific voxeme geometries, which are given below the markup example. In cases where a representation instance is given independently of a geometry, it should be assumed to denote a prototypical or "default" representation of the voxeme's associated lexeme. We explore the richness of the language in more detail in the long version of the paper.

### 4.1. Objects

In this section, we illustrate the visual object concept modeling capabilities for objects by differentiating between properties of the object's type, habitat, afforfance structure, and how it is embodied. Consider the voxeme structures for *wall* and *table*.

$$
(3)\quad \left[\begin{array}{l}
\textbf{wall}\\
\text{LEX} = \left[\begin{array}{l}\text{PRED} = \textbf{wall}\\ \text{TYPE} = \textbf{physobj}\end{array}\right]\\
\text{TYPE} = \left[\begin{array}{l}\text{HEAD} = \textbf{rectangular_prism}\\ \text{COMPONENTS} = \textbf{nil}\\ \text{CONCAVITY} = \textbf{flat}\\ \text{ROTATSYM} = \{X,Y,Z\}\\ \text{REFLECTSYM} = \{XY,XZ,YZ\}\end{array}\right]\\
\text{HABITAT} = \left[\begin{array}{l}\text{INTR} = \left[\begin{array}{l}\text{UP} = align(Y,\mathcal{E}_Y)\\ \text{FRONT} = front(+Z)\\ \text{CONSTR} = Z \ll Y, Z \ll X\end{array}\right]\\ \text{EXTR} = \ldots\end{array}\right]\\
\text{AFFORD_STR} = \left[\begin{array}{l}A_1 = H_1 \rightarrow [E_1]R\\ A_2 = \textbf{...}\\ A_3 = \textbf{...}\end{array}\right]\\
\text{EMBODIMENT} = \left[\begin{array}{l}\text{SCALE} = >\textbf{agent}\\ \text{MOVABLE} = \textbf{false}\end{array}\right]
\end{array}\right]
$$

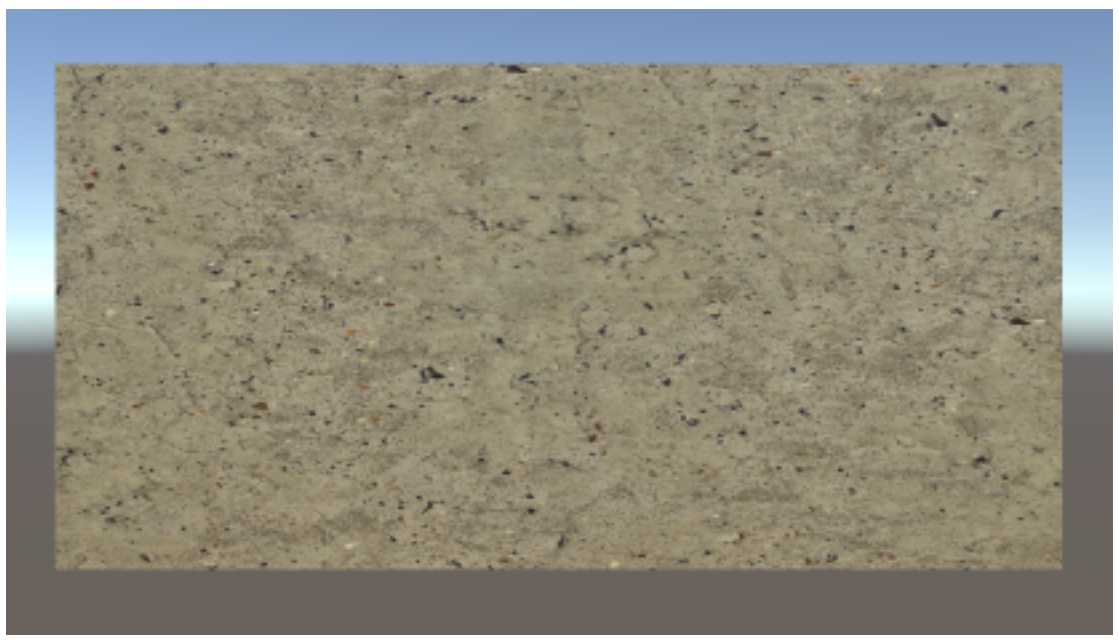

Figure 1: Wall voxeme instance

While walls and tables are both geometries that have habitats which are upwardly aligned, tables have a head geometry of `sheet`, which is a special case of `rectangular_prism` where the Y dimension is significantly less than X or Z. This head is identified in the habitat as the top of the object, facing up.

$$
(4)\quad \left[\begin{array}{l}
\textbf{table}\\
\text{LEX} = \left[\begin{array}{l}\text{PRED} = \textbf{table}\\ \text{TYPE} = \textbf{physobj}\end{array}\right]\\
\text{TYPE} = \left[\begin{array}{l}\text{HEAD} = \textbf{sheet [1]}\\ \text{COMPONENTS} = \textbf{surface [1], leg+}\\ \text{CONCAVITY} = \textbf{flat}\\ \text{ROTATSYM} = \{Y\}\\ \text{REFLECTSYM} = \{XY,YZ\}\end{array}\right]\\
\text{HABITAT} = \left[\begin{array}{l}\text{INTR} = \left[\begin{array}{l}\text{UP} = align(Y,\mathcal{E}_Y)\\ \text{TOP} = top(+Y)\end{array}\right]\\ \text{EXTR} = \ldots\end{array}\right]\\
\text{AFFORD_STR} = \left[\begin{array}{l}A_1 = H_1 \rightarrow [E_1]R\\ A_2 = \textbf{...}\\ A_3 = \textbf{...}\end{array}\right]\\
\text{EMBODIMENT} = \left[\begin{array}{l}\text{SCALE} = \textbf{agent}\\ \text{MOVABLE} = \textbf{true}\end{array}\right]
\end{array}\right]
$$

Now consider the voxeme for *plate*. This illustrates how the habitat feeds into activating the affordance structure associated with the object. Namely, if the appropriate conditions are satisfied ([1]), then the telic affordance associated with a plate is activated; every putting of *x* on *y* results in *y* holding *x*.

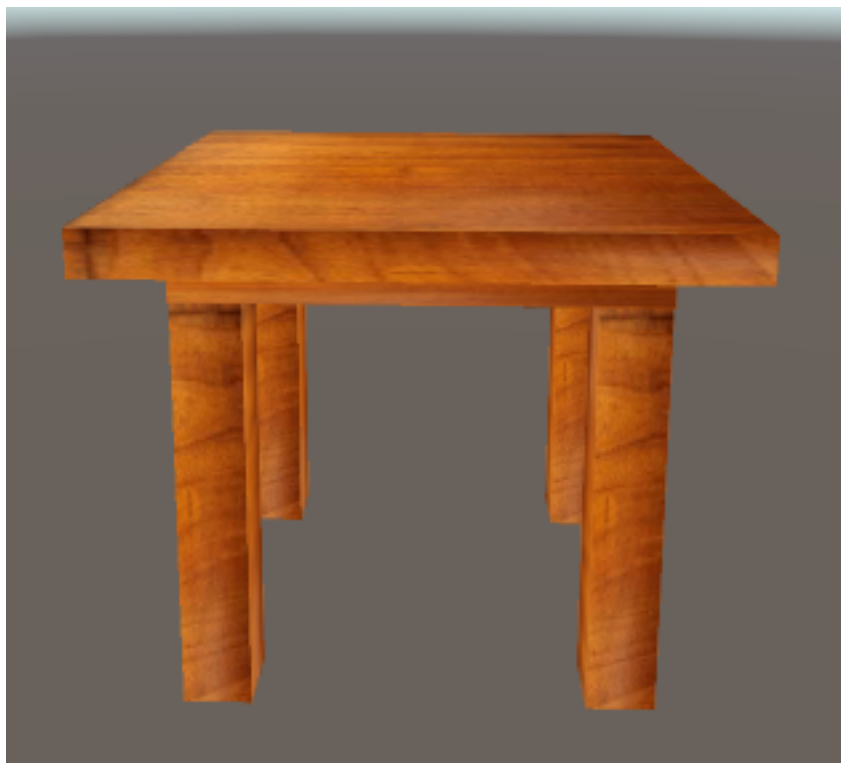

Figure 2: Table voxeme instance

$$
(5)\quad \left[\begin{array}{l}
\textbf{plate}\\
\text{LEX} = \left[\begin{array}{l}\text{PRED} = \textbf{plate}\\ \text{TYPE} = \textbf{physobj}\end{array}\right]\\
\text{TYPE} = \left[\begin{array}{l}\text{HEAD} = \textbf{sheet}\\ \text{COMPONENTS} = \textbf{surface, base}\\ \text{CONCAVITY} = \textbf{concave}\\ \text{ROTATSYM} = \{Y\}\\ \text{REFLECTSYM} = \{XY,YZ\}\end{array}\right]\\
\text{HABITAT} = \left[\begin{array}{l}\text{INTR} = [1]\left[\begin{array}{l}\text{UP} = align(Y,\mathcal{E}_Y)\\ \text{TOP} = top(+Y)\end{array}\right]\\ \text{EXTR} = \ldots\end{array}\right]\\
\text{AFFORD_STR} = \left[\begin{array}{l}A_1 = H[1] \rightarrow [put(x,y)]hold(y,x)\\ A_2 = \textbf{...}\\ A_3 = \textbf{...}\end{array}\right]\\
\text{EMBODIMENT} = \left[\begin{array}{l}\text{SCALE} = <\textbf{agent}\\ \text{MOVABLE} = \textbf{true}\end{array}\right]
\end{array}\right]
$$

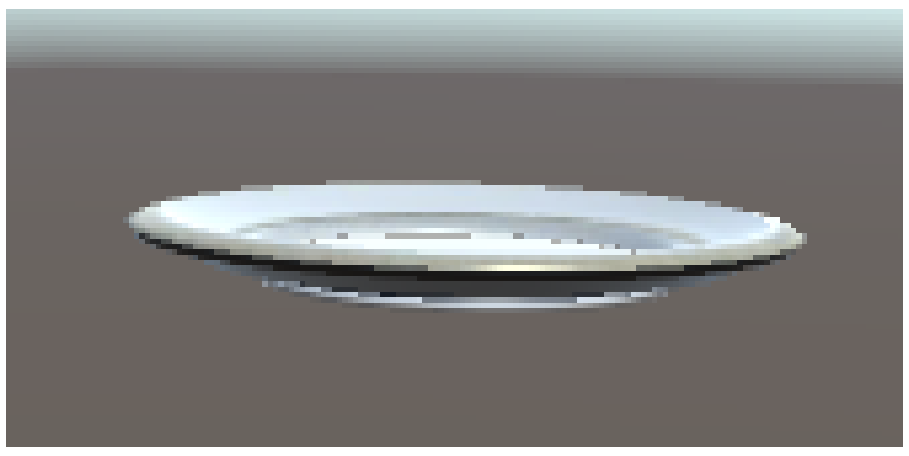

Figure 3: Plate voxeme instance

$$
(6)\quad \left[\begin{array}{l}
\textbf{apple}\\
\text{LEX} = \left[\begin{array}{l}\text{PRED} = \textbf{apple}\\ \text{TYPE} = \textbf{physobj}\end{array}\right]\\
\text{TYPE} = \left[\begin{array}{l}\text{HEAD} = \textbf{ellipsoid [1]}\\ \text{COMPONENTS} = \textbf{fruit [1], stem, leaf}\\ \text{CONCAVITY} = \textbf{convex}\\ \text{ROTATSYM} = \{Y\}\\ \text{REFLECTSYM} = \{XY,YZ\}\end{array}\right]\\
\text{HABITAT} = \left[\begin{array}{l}\text{INTR} = \{\}\\ \text{EXTR} = \ldots\end{array}\right]\\
\text{AFFORD_STR} = \left[\begin{array}{l}A_1 = H_1 \rightarrow [E_1]R\\ A_2 = \textbf{...}\\ A_3 = \textbf{...}\end{array}\right]\\
\text{EMBODIMENT} = \left[\begin{array}{l}\text{SCALE} = <\textbf{agent}\\ \text{MOVABLE} = \textbf{true}\end{array}\right]
\end{array}\right]
$$

## 4.2. Programs

Events are interpreted as programs, moving an object or changing an object property or relation from state to state, as described in more detail in the next section. Program structure derives largely from the lexical semantics of the verb in GL. However, the semantics of the predicative change over the event structure is interpreted operationally. We illustrate this with two predicates as programs, *slide* and *put*.

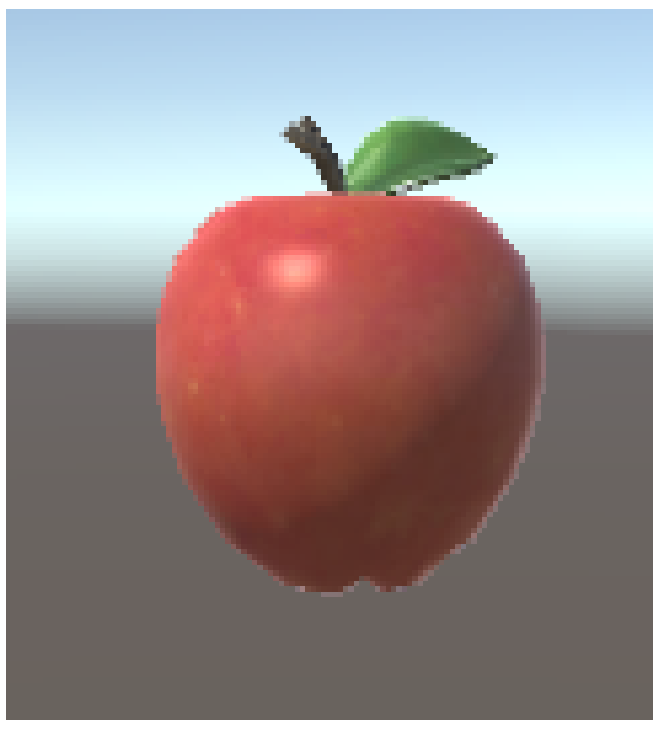

Figure 4: Apple voxeme instance

$$(7)\quad \left[\begin{array}{l} \textbf{slide} \\ \text{LEX} = \left[\begin{array}{l} \text{PRED} = \textbf{slide} \\ \text{TYPE} = \textbf{process} \end{array}\right] \\ \text{TYPE} = \left[\begin{array}{l} \text{HEAD} = \textbf{process} \\ \text{ARGS} = \left[\begin{array}{l} \text{A}_1 = \textbf{x:physobj} \\ \text{A}_2 = \textbf{y:physobj} \end{array}\right] \\ \text{BODY} = \left[\begin{array}{l} \text{E}_1 = while(EC(x,y), \\ \quad move(x)) \end{array}\right] \end{array}\right] \end{array}\right]$$

$$(8)\quad \left[\begin{array}{l} \textbf{put} \\ \text{LEX} = \left[\begin{array}{l} \text{PRED} = \textbf{put} \\ \text{TYPE} = \textbf{transition_event} \end{array}\right] \\ \text{TYPE} = \left[\begin{array}{l} \text{HEAD} = \textbf{transition} \\ \text{ARGS} = \left[\begin{array}{l} \text{A}_1 = \textbf{x:agent} \\ \text{A}_2 = \textbf{y:physobj} \\ \text{A}_3 = \textbf{z:location} \end{array}\right] \\ \text{BODY} = \left[\begin{array}{l} \text{E}_1 = grasp(x,y) \\ \text{E}_2 = [while( \\ \quad hold(x,y), \\ \quad move(x)] \\ \text{E}_3 = [at(y,z) \rightarrow \\ \quad ungrasp(x,y)] \end{array}\right] \end{array}\right] \end{array}\right]$$

### 4.3. Attributes

Unlike physical objects, which can be associated with specific geometries, attributes are abstract predications over distinct domains, and can only be simulated by application to an element of this domain. Below is an example of the nominal attributive interpretation of the adjective *brown*, and the ordinal attributive interpretation of *small*.

$$(9)\quad \left[\begin{array}{l} \textbf{brown} \\ \text{LEX} = \left[\begin{array}{l} \text{PRED} = \textbf{brown} \end{array}\right] \\ \text{TYPE} = \left[\begin{array}{l} \text{SCALE} = \textbf{nominal} \\ \text{ARITY} = \textbf{intransitive} \\ \text{ARG} = \textbf{x:physobj} \end{array}\right] \end{array}\right]$$

$$(10)\quad \left[\begin{array}{l} \textbf{small} \\ \text{LEX} = \left[\begin{array}{l} \text{PRED} = \textbf{small} \end{array}\right] \\ \text{TYPE} = \left[\begin{array}{l} \text{SCALE} = \textbf{ordinal} \\ \text{ARITY} = \textbf{transitive} \\ \text{ARG} = \textbf{x:physobj} \end{array}\right] \end{array}\right]$$

### 4.4. Relations

Spatial relations are propositional expressions, contributing configurational information about two or more objects in a state.

$$(11)\quad \left[\begin{array}{l} \textbf{touching} \\ \text{LEX} = \left[\begin{array}{l} \text{PRED} = \textbf{is_touching} \end{array}\right] \\ \text{TYPE} = \left[\begin{array}{l} \text{CLASS} = \textbf{config} \\ \text{VALUE} = \textbf{EC} \\ \text{ARGS} = \left[\begin{array}{l} \text{A}_1 = \textbf{x:3D} \\ \text{A}_2 = \textbf{y:3D} \\ \text{A}_3 = \textbf{...} \end{array}\right] \end{array}\right] \end{array}\right]$$

### 4.5. Functions

Finally, we illustrate the semantics of spatial functions with *top*. This applies to an object of dimensionality $n$, returning an object of dimensionality $n - 1$; the top of a cube is a plane; that of a rectangular sheet is a line; and the top of a line is a point.

$$(12)\quad \left[\begin{array}{l} \textbf{top} \\ \text{LEX} = \left[\begin{array}{l} \text{PRED} = \textbf{top} \end{array}\right] \\ \text{TYPE} = \left[\begin{array}{l} \text{ARG} = \textbf{x:physobj} \\ \text{REFERENT} = \textbf{x}\rightarrow\text{HEAD} \\ \text{MAPPING} = \textbf{dimension(n):n-1} \\ \text{ORIENTATION} = \left[\begin{array}{l} \text{SPACE} = \textbf{world} \\ \text{AXIS} = \textbf{+Y} \\ \text{ARITY} = \textbf{x}\rightarrow\text{HABITAT}\rightarrow \\ \quad \text{INTR}[top(axis)]: \\ \quad intransitive \end{array}\right] \end{array}\right] \end{array}\right]$$

## 5. Using VoxML for Simulation Modeling of Language

VoxML treats objects and events in terms of a dynamic event semantics, Dynamic Interval Temporal Logic (DITL) (Pustejovsky and Moszkowicz, 2011). The advantage of adopting a dynamic interpretation of events is that we can map linguistic expressions directly into simulations through an operational semantics (Miller and Charles, 1991; Miller and Johnson-Laird, 1976). Models of processes using updating typically make reference to the notion of a state transition (van Benthem, 1991; Harel, 1984). This is done by distinguishing between formulae, $\phi$, and programs, $\pi$. A formula is interpreted as a classical propositional expression, with assignment of a truth value in a specific model. For the present discussion, we represent the dynamics of actions in terms of Labeled Transition Systems (LTSs) (van Benthem, 1991).[3] An LTS consists of a triple, $\langle S, Act, \rightarrow\rangle$, where: $S$ is the set of states; $Act$ is a set of actions; and $\rightarrow$ is a total transition relation: $\rightarrow \subseteq S \times Act \times S$. An action, $\alpha \in Act$, provides the labeling on an arrow, making it explicit what brings about a state-to-state transition. As a shorthand for $(e_1, \alpha, e_2) \in \rightarrow$, we will also use $e_1 \xrightarrow{\alpha} e_2$. Simulation generation begins by parsing a natural English sentence, currently using the ClearNLP parser (Choi and McCallum, 2013).[4] The dependency parse is then transformed into a predicate-argument set representation using the root of the parse as the main predicate and its dependencies as arguments. Each predicate can have more than one argument and arguments may themselves be predicates (thus higher-order predicates are permissible). This predicate-argument set formula is then evaluated from the innermost first order predicates outward until a single first-order representation is reached.

[3]This is consistent with the approach developed in (Fernando, 2009; Fernando, 2013). This approach to a dynamic interpretation of change in language semantics is also inspired by Steedman (2002).

[4]We are working toward integrating the lexical entries with a rule-based parser that provides for a compositionally derived interpretation of the sentence being parsed.

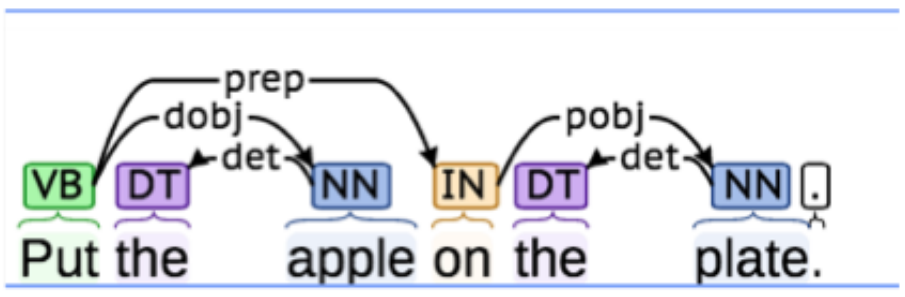

Figure 5: Dependency parse

| 1. *pred* := *put(x,y)* |
| --- |
| 2. *x* := *apple* |
| 3. *y* := *on(z)* |
| 4. *z* := *plate* |
| *put(apple,on(plate))* |

Figure 6: Transformation to logical form

*put(apple,<1, 2.3, -0.8>)*

Figure 7: "on(plate)" evaluates to coordinates

The evaluation of predicates entails the composition of the voxemes involved. Since we allow for program voxemes (verbs), logical type voxemes (relations, attributes, functions) and object voxemes (nominals), evaluation involves executing a snippet of code that operationalizes a program voxeme using the geometric and VoxML-encoded semantic information from the voxeme(s) indicated by the input predicate's argument(s).

In the example given above, the predicate *on(plate)* is evaluated to a specific location that satisfies the constraint denoted by "on the plate", using information about the *plate* object, specifically its dimensions and concavity. The program *put* can then be realized as a DITL program that moves the object *apple* toward that location until the apple's location and the location evaluated for *on(plate)* are equal, and executed as a simple state transition $\boxed{\neg\phi}_{e_1} \xrightarrow{\alpha} \boxed{\phi}_{e_2}$.

Given a 3D scene that contains voxemes for all nominals referred to in the text, the program can be operationalized in code and the state transition can be executed over the geometries visualized in 3D space. We use an updated version of the the simulator built for (Pustejovsky and Krishnaswamy, 2014) and the C# language to generate visualizations like that shown in Figure 9 below.

## 6. Conclusion

In this paper we have outlined a specification to represent semantic knowledge of real-world objects represented as three-dimensional models. We use a combination of parameters that can be determined from the object's geometrical properties as well as lexical information from natural language, with methods of correlating the two where applicable. This information allows for the visualization and simulation software to fill in information missing from the natural language input and allows the software to render a functional visualization of programs being run over objects in a robust and extensible way. Currently we have a voxicon containing 500 object (noun) voxemes and 10 program (verb) voxemes.

| *put(obj,y)* |
| --- |
| *loc(obj) := x, target(obj) := y; b := x;* <br> $(x := w;\ x \neq w;\ d(b,x) < d(b,w),\ d(b,y) > d(y,w)^{+}$ |

Figure 8: DITL expressions for *put(obj,y)*.

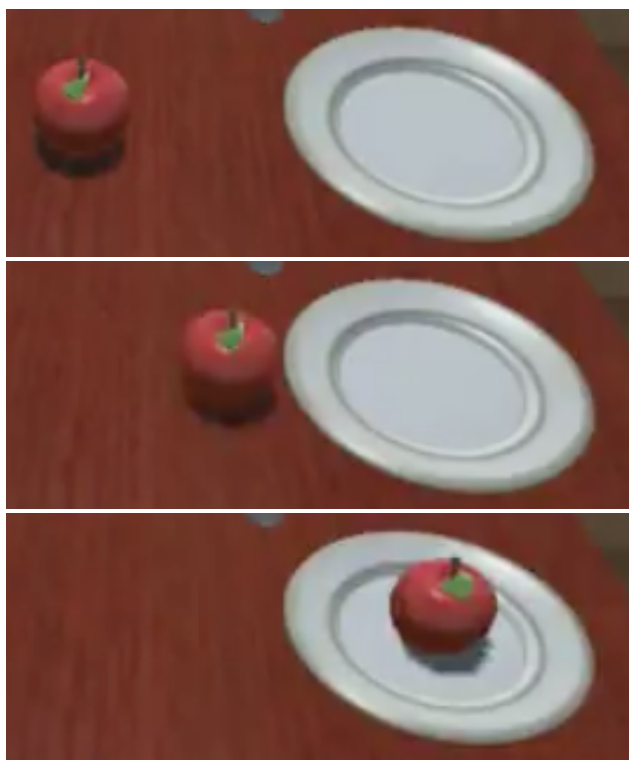

Figure 9: Program is executed in automatically generated rendering

As our library of available voxemes continues to grow the specification elements may as well, allowing us to operationalize a larger library of various and more complicated programs, and to compose complex behaviors out of simpler ones. The voxeme library and visualization software will be deployed at http://www.voxicon.net. There users may find the current library of voxemes and conduct visualizations of available behaviors driven by VoxML after parsing and interpretation.

## 7. Acknowledgements

This work was supported by Contract W911NF-15-C-0238 with the US Defense Advanced Research Projects Agency (DARPA) and the Army Research Office (ARO). Approved for Public Release, Distribution Unlimited. The views expressed are those of the authors and do not reflect the official policy or position of the Department of Defense or the U.S. Government. All errors and mistakes are, of course, the responsibilities of the authors.

## 8. Bibliographical References